\documentclass{article}

     \usepackage[preprint]{neurips_2020}

\usepackage[utf8]{inputenc} 
\usepackage[T1]{fontenc}    
\usepackage{hyperref}       
\usepackage{url}            
\usepackage{booktabs}       
\usepackage{amsfonts}       
\usepackage{nicefrac}       
\usepackage{microtype}      

\usepackage{amsmath}
\usepackage{amsfonts}
\usepackage{amssymb}
\usepackage{natbib}
\usepackage{graphicx}
\usepackage{geometry}
\usepackage{xcolor}
\usepackage{mathtools}
\usepackage{tikz}
\usetikzlibrary{arrows}
\usepackage{tikz-qtree}
\usetikzlibrary{trees}
\usetikzlibrary{automata,positioning}
\usepackage{fontawesome}
\usepackage{caption}
\usepackage{subcaption}
\usepackage{multirow}
\usepackage{enumitem}

\newcommand\ci{\perp\!\!\!\perp}
\newcommand{\E}{\mathbb{E}}
\newcommand{\G}{{\mathcal G}}
\DeclareMathOperator{\pa}{pa}

\title{Causal Inference in the Presence of Interference in Sponsored Search Advertising}

\author{
	Razieh Nabi\thanks{Work done during internship at Microsoft Research.}  \\ Johns Hopkins University \\ rnabi@jhu.edu
	\And 
	Joel Pfeiffer  \\ Microsoft Bing Ads \\ joelpf@microsoft.com
	\And 
	Murat Ali Bayir  \\ Microsoft Bing Ads \\ mbayir@microsoft.com  
	\And 
	Denis Charles \\ Microsoft Bing Ads \\ cdx@microsoft.com
	\And 
	Emre  K\i c\i man  \\ Microsoft Research  \\ emrek@microsoft.com
}

\begin{document}

\maketitle

\begin{abstract}
	In classical causal inference, inferring cause-effect relations from data  relies on the assumption that units are independent and identically distributed. This assumption is  violated in settings where units are related through a network of dependencies. An example of such a setting is ad placement in sponsored search advertising, where the clickability of a particular ad is potentially influenced by where it is placed and where other ads are placed on the search result page. In such scenarios, confounding arises due to not only the individual ad-level covariates but also the placements and covariates of other ads in the system. In this paper, we leverage the language of causal inference in the presence of interference to model interactions among the ads. Quantification of such interactions allows us to better understand the click behavior of users, which in turn impacts the revenue of the host search engine and enhances user satisfaction. We illustrate the utility of our formalization through experiments carried out on the ad placement system of the Bing search engine. 
\end{abstract}

\section{Introduction}
\label{sec:intro}

In recent years, advertisers have increasingly shifted their ad expenditures online. One of the most effective platforms for online advertising is search engine result pages. Given a user query, the search engine allocates a few ad slots (e.g., above or below its organic search results) and runs an auction among advertisers who are bidding and competing for these slots. Quantifying the effectiveness of ad placement is vital not only to the experience of the user, but also revenue of the advertiser and the search engine. Click yield is a common metric used in this regard. Often, statistical models are used to predict the click behavior of users by estimating the likelihood of receiving a click in a given slot using logged data. A rich literature is devoted to click prediction in sponsored search advertising \citep{xiong2012relational, cheng2012multimedia, nabi2015conversion, zhang2014sequential, effendi2017click}. However, a comprehensive understanding of click behavior requires causal, rather than associative, reasoning \citep{bottou2013counterfactual, hill2015measuring, zeng2019causal, yin2014estimating}. 

Causal inference is central to making data-driven decisions.
Inferring valid cause-effect relations, even with granular data and large sample sizes, is complicated 
by 
confounding induced by common causes of observed exposures and outcomes. In classical causal inference, it is assumed that samples are \textit{independent and identically distributed} (iid). However, a causal view of ad placement under the iid assumption is implausible as ads interfere with one another from the beginning of the auction until the end when clicks on impressed ads are recorded. 
In non-iid settings, confounding arises due to not only the individual ad-level covariates but also the exposures and covariates of other ads in the system. 
This is commonly referred to as \textit{interference} \citep{hudgens2008toward}. 
Incorporating knowledge of interference into the statistical models used to compute rank scores for each ad can help optimize the profitability of the final layout of each search page. Moreover,  a proper understanding of the interference issue in relation to causal inference directly impacts engineering of more purposeful interventions and design of more effective A/B testing for ad placement. 


In this paper, we formalize the problem of interference among ads using the language of causal inference. To the best of our knowledge, this is the first attempt to analyze ads under the plausible and realistic setting of interference. Throughout the paper, we discuss mechanisms that give rise to interference in ad placement. Using graphical models, we assume a causal structure that encodes the various sources of interference. We formulate our causal questions and discuss the identification and estimation of relevant  effects. Our experiments find statistically significant interference effects among ads.   
We further 
adapt the \textit{constraint-based} structure learning algorithm  \textit{Fast Causal Inference} \citep{spirtes2000causation} 
to verify the soundness of our presumed causal structure and 
learn the underlying mechanisms that give rise to interference. 
Finally, 
we incorporate the knowledge of interference 
to improve the performance of the statistical models used during the course of the auction. We demonstrate this improvement in performance by running experiments that closely resemble the framework in the  Genie model -- an offline counterfactual policy estimation framework for optimizing Sponsored Search Marketplace in Bing ads \citep{bayir2019genie}. 


\section{Preliminaries and Setup}
\label{sec:prelim}

In causal inference, we are interested in quantifying the cause-effect relationships between a treatment variable $A$ and an outcome $Y$ using experimental or observational data. A common setting assumes that the treatment received by one unit does not affect the outcomes of other units -- this is known as the \textit{stable unit treatment value assumption} or SUTVA \cite{rubin1980randomization} and is informally referred to as the ``no-interference'' assumption. In this setting, the \emph{average causal effect (ACE)} of a binary treatment $A$ on $Y$ is defined as $\textit{ACE} \coloneqq \E[Y(1)] - \E[Y(0)],$ where $Y(a)$ denotes the counterfactual/potential outcome $Y$ had treatment $A$ been assigned to $a,$ possibly contrary to the fact.    

Causal inference uses assumptions in causal models to link the observed data distribution to the distribution over counterfactual random variables. A simple example of a causal model is the \textit{conditionally ignorable model} which encodes three main assumptions: 
(i) \textit{Consistency} assumes the mechanism that determines the value of the outcome does not distinguish the method by which the treatment was assigned, as long as the treatment value assigned was invariant, 
(ii) \textit{Conditional ignorability} assumes $Y(a) \perp A \mid X,$ where $X$ acts as a set of observed confounders, such that adjusting for their influence suffices to remove all non-causal dependence between $A$ and $Y,$ and 
(iii) \textit{Positivity} of $p(A = a \mid X = x), \forall a,x.$ Under these assumptions, $p(Y(a))$ is identified as the following function of the observed data: {\small $\sum_X p(Y \mid A = a, X) \times p(X),$} known as \textit{backdoor adjustment} or \textit{g-formula} \citep{pearl2009causality, robins1986new}. For a general identification theory of causal effects in the presence of unmeasured confounders see \citep{tian2002general, shpitser2006identification, huang2006pearl, bhattacharya2020semiparametric}. 
Alternative causal quantities of interest include conditional causal effects (effects within subpopulations defined by covariates)\cite{shpitser2012identification}, mediation quantities (which decompose effects into components along different mechanisms)\cite{shpitser2013counterfactual}, and the effects of decision rules in sequential settings (such as dynamic treatment regimes in personalized medicine)\cite{nabi2018estimation}.


In this paper, we relax the implausible assumption of no-interference in ad placement. 
Interference among ads across different pageviews 
creates the most extreme scenario of \textit{full interference,} as this allows for user interaction with the system over multiple time frames.  
Following the convention in \citep{sobel2006randomized, hudgens2008toward, tchetgen2012causal, ogburn2014causal}, we model only interference within pageviews and restrict any cross-pageview interference among ads. In other words, we restrict the interference to spatial constraints and exclude temporal dependence across pageviews. This is known as \textit{partial interference} and could be justified by the fact that pageviews are query specific and are separated by time and space. In presence of interference, the counterfactual $Y(a)$ is no longer well-defined as we need to distinguish ads by a proper indexing scheme and consider the treatment assignments of other ads simultaneously. 

Suppose we have $N$ pageviews, indexed by $n = 1, \ldots, N,$ with each containing $m$ impressed ads. We index the ads on each pageview by $i = 1, \dots, m$ based on the order in which they appear on the page. The $i$-th ad on the $n$-th pageview is represented by the tuple $(X_{ni}, A_{ni}, Y_{ni}),$ where $X_{ni}$ denotes the vector that collects all the ad-specific features such as geometric features (e.g., line width, pixel height), decorative features (e.g., rating information, twitter followers), and other textual features extracted from the ad. $ A_{ni}$ denotes the treatment and is predefined by the analyst. An example of a treatment is the \textit{block} membership of the ad: an indicator that specifies whether the ad is placed on top of the page (Top) or bottom of the page (Bottom). Ads can also appear elsewhere such as the sidebars. In this paper, without loss of generality, we assume we only have two distinct blocks of ads on each pageview: Top and Bottom. $Y_{ni}$ denotes a binary indicator of receiving a click by the user. We denote the state space of a random variable $V$ by $\mathfrak{X}_V.$

Let ${\bf X}_n \coloneqq (X_{n1}, \dots, X_{nm}),$ ${\bf A}_n \coloneqq (A_{n1}, \dots, A_{nm}),$ and ${\bf Y}_n \coloneqq (Y_{n1}, \dots, Y_{nm})$ collect the  features, treatment assignments, and outcomes  of all the ads on the $n$-th pageview, respectively. We define the counterfactual $Y_{ni}({\bf a}_n)$ to be the click response of the $i$-th ad on the $n$-th pageview where every ad on the same pageview is relocated according to the treatment assignment rule ${\bf a}_n,$ which is a vector of size $m$ and the $i$-th element $a_i$ denotes the treatment value of the $i$-th ad.  This notation makes the interference among ads on the same pageview more explicit as the potential outcome of a single ad now depends on the entire treatment assignment ${\bf a}_n,$ rather than just $a_{ni}.$ The causal effect of interventions in the presence of interference can be quantified by comparing such counterfactuals under different interventions; for instance $Y_{ni}({\bf a}_n)$ vs $Y_{ni}({\bf a'}_n),$ where ${\bf a}_n$ and ${\bf a'}_n$ denote two plausible interventions. 

In the next section, we discuss various sources that give rise to interference among ads and propose a causal graphical model that captures such interactions in a reasonable way. In what follows, we discuss various ways of quantifying the interference effect among ads and provide sufficient conditions for identification of such effects along with estimation strategies. 


\section{Ad Placement in the Presence of Interference}
\label{sec:ad_model}

We describe ad placement in the presence of interference by a system of nonparametric structural equation models with independent errors \citep{pearl2009causality}. 
The key characteristic of structural models is that they represent each variable as deterministic functions of their direct causes together with an unobserved exogenous noise term, which itself represents all causes outside of the model. 
%
Let $U$ denote a variable capturing user intention which is unknown and hidden to the analyst. 
Given such intent, the user types a query, denoted by $C,$ which is expressed as an unrestricted function of the intent $U$ and a noise term $\epsilon_c,$ denoted by $f_c(.).$ 
Upon observing the query, a set of ads are selected from the inventory, then online auction is run to determine winner ads to be displayed on the page. 
The $i$-th displayed ad is denoted by $X_i.$ The relation between $X_i$ and $C$ is captured by an unrestricted function $f_{x_i}(.)$ and the perturbation term $\epsilon_{x_i}.$ The block allocation of $i$-th ad is denoted by $A_i.$ The set of all impressed ads and the allocations are denoted by $\bf X$ and $\bf A,$ respectively (we suppress the indexing of pageviews for clarity.)  The information on $U, {\bf X}, {\bf A},$ along with the noise term $\epsilon_{y_i},$ determines whether the $i$-th ad is clicked or not which is captured by $Y_i.$ The structural equation models are summarized as follows.
{\small
	\begin{align}
	&\textit{user intent }  &  U \ &\leftarrow \ f_u(\epsilon_u) 
	\nonumber \\
	&\textit{user query }  & C \ &\leftarrow \ f_c(U, \epsilon_c)  \nonumber \\ 
	&\textit{i-th ad} &  X_i \ &\leftarrow \ f_{x_i}(C, \epsilon_{x_i})  \nonumber \\ 
	&\textit{i-th ad's allocation} & A_i \ &\leftarrow \ f_{a_i}({\bf X}, \epsilon_{a_i}) \nonumber \\ 
	&\textit{i-th ad's click indication} & Y_i \ &\leftarrow \ f_{y_i}(U, {\bf X}, {\bf A}, \epsilon_{y_i}) \label{eq:sem}  
	\end{align}
}

\vspace{-0.3cm}
 Note that in the above equations, when allocating the $i$-th ad to Top or Bottom, we are not only considering the corresponding features of the ad itself, but also features of other ads on the page, hence the entire array of $\bf X$ is acting as causes of $A_i.$ Similarly, we allow for the entire vector of $\bf A$ and array of $\bf X$ to influence $Y_i.$ These equation capture the interference mechanism in ad placement. In the absence of interference, the above equations simplify by replacing the allocation structural equation with $A_i \leftarrow f_{a_i}(X_i, \epsilon_{a_i})$ and the click indication structural equation with $Y_i \leftarrow f_{y_i}(U, X_i, A_i, \epsilon_{y_i}).$

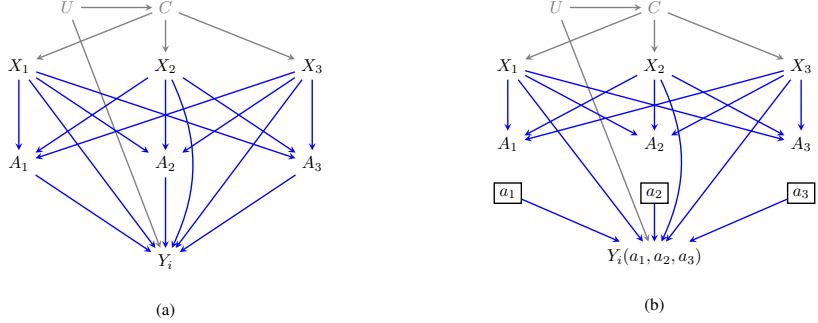
\begin{figure}[!t]
	\begin{center}
		\scalebox{0.65}{
			\begin{tikzpicture}[>=stealth, node distance=2cm]
			\tikzstyle{format} = [minimum size=1.0mm, inner sep=1pt]
			\tikzstyle{square} = [draw, thick, black, minimum size=1.0mm, inner sep=3pt]
			\begin{scope}[xshift=0cm, yshift=0cm]
			\path[->, thick]
			node[] (u) {\textcolor{gray}{$U$}}
			node[right of=u, xshift=-0.0cm, yshift=0.0cm] (c) {\textcolor{gray}{$C$}}
			
			node[below of=c, xshift=0cm, yshift=0.8cm] (x2) {$X_2$}
			node[below of=c, xshift=-3cm, yshift=0.8cm] (x1) {$X_1$}
			node[below of=c, xshift=3cm, yshift=0.8cm] (x3) {$X_3$}
			
			node[below of=x1, yshift=-0.0cm] (a1) {$A_1$}
			node[below of=x2, yshift=-0.0cm] (a2) {$A_2$}
			node[below of=x3, xshift=0.0cm] (a3) {$A_3$}
		
			node[below of=a2, yshift=-0.0cm] (y2) {$Y_i$}
			
			(u) edge[gray] (c)
			(u) edge[gray] (y2)
			
			(c) edge[gray] (x1)
			(c) edge[gray] (x2)
			(c) edge[gray] (x3)
			
			(x1) edge[blue] (a1)
			(x1) edge[blue] (a2)
			(x1) edge[blue] (y2)
			(x1) edge[blue] (a3)
			
			(x2) edge[blue] (a2)
			(x2) edge[blue, bend left=25] (y2)
			(x2) edge[blue] (a1)
			(x2) edge[blue] (a3)
			
			(x3) edge[blue] (a3)
			(x3) edge[blue] (a1)
			(x3) edge[blue] (a2)
			(x3) edge[blue] (y2)
			
			(a1) edge[blue] (y2)
			(a3) edge[blue] (y2)
			
			(a2) edge[blue] (y2)
			
			node[below of=y2, xshift=0cm, yshift=1cm] {(a)};
			\end{scope}
			\begin{scope}[xshift=10cm, yshift=0cm]
			\path[->, thick]
			node[] (u) {\textcolor{gray}{$U$}}
			node[right of=u, xshift=-0.0cm, yshift=0.0cm] (c) {\textcolor{gray}{$C$}}
			
			node[below of=c, xshift=0cm, yshift=0.8cm] (x2) {$X_2$}
			node[below of=c, xshift=-3cm, yshift=0.8cm] (x1) {$X_1$}
			node[below of=c, xshift=3cm, yshift=0.8cm] (x3) {$X_3$}
			
			node[below of=x1, yshift=0.4cm] (a1) {$A_1$}
			node[below of=x2, yshift=0.4cm] (a2) {$A_2$}
			node[below of=x3, yshift=0.4cm] (a3) {$A_3$}
			node[square, below of=a1, yshift=1.cm] (a1_0) {$a_1$}
			node[square, below of=a2, yshift=1.cm] (a2_0) {$a_2$}
			node[square, below of=a3, yshift=1.cm] (a3_0) {$a_3$}
			
			node[below of=a2, yshift=-0.3cm] (y2) {$Y_i(a_1, a_2, a_3)$}
			
			(u) edge[gray] (c)
			(u) edge[gray] (y2)
			
			(c) edge[gray] (x1)
			(c) edge[gray] (x2)
			(c) edge[gray] (x3)
			
			(x1) edge[blue] (a1)
			(x1) edge[blue] (a2)
			(x1) edge[blue] (y2)
			(x1) edge[blue] (a3)
			
			(x2) edge[blue] (a2)
			(x2) edge[blue, bend left=25] (y2)
			(x2) edge[blue] (a1)
			(x2) edge[blue] (a3)
			
			(x3) edge[blue] (a3)
			(x3) edge[blue] (a1)
			(x3) edge[blue] (a2)
			(x3) edge[blue] (y2)
			
			(a1_0) edge[blue] (y2)
			(a3_0) edge[blue] (y2)
			(a2_0) edge[blue] (y2)
			
			node[below of=y2, xshift=0cm, yshift=1cm] {(b)};
			\end{scope}
			\end{tikzpicture}
		}
		\caption{(a) DAG representation of the SEM in \ref{eq:sem} for a pageview with three impressed ads (the independent error terms are omitted from the graph for simplicity.) (b) The corresponding SWIG where we intervene on $\bf A$ and set the block allocations $(A_1, A_2, A_3)$ to $(a_1, a_2, a_3).$}
		\label{fig:npsem}
	\end{center}
\end{figure}

Causal relationships are often represented by graphical causal models \citep{spirtes2000causation, pearl2009causality}. Such models generalize independence models on directed acyclic graphs (DAGs) to also encode conditional independencies on counterfactual variables \citep{richardson2013single}. A DAG $\G(V)$ consists of a set of nodes $V$ connected through directed edges such that there are no directed cycles. We will abbreviate $\G(V)$ as simply $\G,$ when the vertex set is clear from the given context. Statistical models of a DAG $\G$ are sets of distributions that factorize as $p(V) = \prod_{V_i \in V} p(V_i \mid \pa_\G(V_i))$, where $\pa_\G(V_i)$ are the parents of $V_i$ in $\G$. The absence of edges between variables in $\G,$ relative to a complete DAG entails conditional independence facts in $p(V).$ These can be directly read off from the DAG $\G$ by the well-known d-separation criterion \citep{pearl2009causality}. That is, for disjoint sets $X, Y, Z$, the following \emph{global Markov property} holds: $(X \ci_{\text{d-sep}} Y \mid Z)_\G \implies (X \ci Y \mid Z)_{p(V)}.$ When the context is clear, we will simply use $X \ci Y \mid Z$ to denote the conditional independence between $X$ and $Y$ given $Z.$ The DAG representation of the structural equations in (\ref{eq:sem}) for a pageview with three impressed ads is shown in Fig.~\ref{fig:npsem}(a). For simplicity and to avoid cluttering the graph, we only depict the outcome of the $i$-th ad on the DAG and marginalize out all the other outcomes (since all the outcomes share the same set of parents.) The statistical model of the DAG in Fig.~\ref{fig:npsem}(a), assuming all outcomes are included on the DAG, can be written as,  
{\small
$
p(U, C, {\bf X, A, Y}) = p(U) \times p(C \mid U) \times \prod_{i = 1}^3 \ \big\{ p(X_i \mid C) \times p(A_i \mid {\bf X}) \times p(Y_i \mid U, {\bf X, A}) \big\}. 
$
}

As we mentioned earlier, the user intent is unmeasured. We further restrict our attention to ad-specific features and leave the query-specific features aside. In other words $U$ and $C$ are both treated as latent. We highlight this in Fig~\ref{fig:npsem}(a) by coloring both vertices and the relevant edges in gray.  In this case, the joint distribution over observed variables ${\bf X, A, Y}$ and latent variables $U, C$ is said to be Markov relative to a hidden variable DAG.  There may be infinitely many hidden variable DAGs that imply the same set of conditional independences on the observed margin, i.e., $p({\bf X, A, Y}).$ It is typical to use a single acyclic directed mixed graph that entails the same set of equality constraints as this infinite class; see  \cite{verma1990equivalence, richardson2017nested} for more details. 

\subsection{Sources of Interference in Ad Placement}

In order to better understand the interference behavior among ads, we need to identify the causal mechanisms that give rise to such behaviors. Looking at our causal model in Fig.~\ref{fig:npsem}(a), we allow for two distinct pathways through which other ads influence $Y_i.$ One is direct pathways such as $X_j \rightarrow Y_i$ and $A_j \rightarrow Y_i$. 
This type of interference is called \textit{direct interference}. As an example, suppose a low quality ad (determined by various scores) is placed in the Top.
The poor quality of this ad may shape the user's opinion about 
the sorted search results in negative ways, preventing them from clicking on further ads. Similarly, placing a high quality ad in the Top may convince the user to return and explore more ads. Other pathways by which outcomes of different ads could be related are ones that go through the common unmeasured confounders and account for marginal dependencies between $Y_i$ and $Y_j.$ An example of this marginal dependency is through user intent $U,$ $Y_j \leftarrow U \rightarrow Y_i.$ This type of interference is called \textit{interference by homophily} \citep{shalizi2011homophily}. Accounting for homophily makes our framework more practical as it allows for unmeasured confounders to influence multiple outcomes simultaneously.  

The third type of interference that we account for is called \textit{allocational interference}. In allocational interference, the interactions among units are modeled according to their corresponding group assignments. Through interactions within a group, units' characteristics may affect one another. This type of interference is well-suited for our purposes since each pageview is divided into non-overlapping blocks (Top and Bottom), and we can simply treat each block as a single group of ads. In our setting, treatment allocates each ad to a single block (randomly or given covariates $\bf X$), and the outcome of the ad is affected by which other ads are allocated to the same block. We call this behavior \textit{block-level interference.} 
We can also imagine a scenario where the outcome of an ad is affected by the ads that are {\bf not} allocated to the same block. In other words, ads could potentially interact across blocks. We call this \textit{cross-block interference.} 
As an example, moving a high quality ad to the Bottom may improve the perception of other ads in the Bottom and yield higher clicks on these ads. On the other hand, it may also affect the click yields of ads in the Top by drawing attention away from these ads, resulting in cross-block interactions. 
In order to formalize the block-level interference and cross-block interference, we split ${\bf X}$ into two disjoint sets: one that contains block-level information, denoted by ${\bf X}^b,$ and one that contains information outside the block, denoted by ${\bf X}^c.$ For the $i$-th positioned ad, we define two disjoint sets: 
\begin{align*}
{\bf X}^b_{i} &= \big\{ X_{j} \in {\bf X}  \ \text{ s.t. } \ A_j = A_i \big\} = \big\{ \mathbb{I}(A_j = A_i) \times X_j, \ \forall j = 1, \dots, m \big\},  \\
{\bf X}^c_{i} &= \big\{ X_{k} \in {\bf X} \ \text{ s.t. } \ A_j \not= A_i \big\} =  \big\{ \mathbb{I}(A_j \not= A_i) \times X_j, \ \forall j = 1, \dots, m \big\}.
\end{align*}

We modify the structural equations for $Y_i$ in (\ref{eq:sem}) to directly account for the allocational interference in our framework by simply replacing $f_{y_i}(U, {\bf X, A}, \epsilon_{y_i})$ with $f_{y_i}(U, {\bf X}^b_{i}, {\bf X}^c_{i}, \epsilon_{y_i}).$ Note that both ${\bf X}^b_{i} $ and ${\bf X}^c_{i}$ depend on the treatment rule $\bf A$ by construction. The function $f_{y_i}$ can take a nonlinear or a linear form. For illustration, assume $f_{y_i}$ is linear in parameters. Therefore, we have: 
\begin{align*}
Y_i\ \gets \ \sum_{j = 1}^m \gamma_j \times \mathbb{I}(A_j = A_i) \times X_j +  \eta_j \times \mathbb{I}(A_j \not= A_i) \times X_j + \epsilon_{y_i} 
\end{align*}
In the above equation, $\gamma_j$ controls the block-level influence of $X_j$ on the $i$-th ad if $X_j$ is in the same block as $X_i,$ otherwise the influence is controlled by the parameter $\eta_j.$ If $\eta_j = 0, \forall j,$ then this implies that there is no cross-block interference and blocks are independent. If $\eta_j = \gamma_j, \forall j,$ then this implies that there is no allocational interference. In other words, interactions within blocks and across blocks are modeled exactly the same and therefore the notion of ``groups'' is ruled out.  


\section{Interference Effects Among Ads}
\label{sec:effects} 

Structural equation models, such as the one in display~(\ref{eq:sem}), enable us to determine the response of variables to interventions through incorporating knowledge of the functional dependencies between variables. For instance, intervening on the block allocation of the $i$-th ad would fix the value of $A_i$ to $a_i,$ and would transform descendants of $A_i$ to counterfactual variables of the form $V(a_i).$
Under an intervention that sets $\bf A$ to $\bf a,$ the structural equations in (\ref{eq:sem}) are modified as follows: 
\begin{align}
A_i \ &\leftarrow \ a_i, \ \forall i = 1, \dots, m, \quad \text{ and } \quad Y_i({\bf a}) \ \leftarrow \ f_{y_i}(U, {\bf X}, {\bf a}, \epsilon_{y_i}),  \ \forall i = 1, \dots, m.
\label{eq:intervention}
\end{align}

Interventions can be directly applied to the causal graph through a node-splitting operation where random variables in $\bf A$ are split into two parts: a random part that takes all the incoming edges and a fixed part that takes all the outgoing edges. The resulting graph is called a single-world intervention graph (SWIG) which encodes counterfactual independences associated with the intervention \citep{richardson2013single}. Given the causal model in Fig.~\ref{fig:npsem}(a), we obtain the corresponding SWIG in Fig.~\ref{fig:npsem}(b) after performing the intervention described in display (\ref{eq:intervention}).

\subsection{Causal Effects of Interest} 

We set block allocation as our treatment of interest, and based on the prior literature, consider several causal effects that are of particular interest in ad placement systems. 

\begin{enumerate}[leftmargin=*]
	\item \textit{Unit-level effect}: defined as the effect of modifying an ad’s block allocation on its clickability but holding the block allocations of other ads fixed. 
	Assume we have a fixed allocation rule ${\bf a},$ and we are interested in moving the $i$-th ad from block $a'$ to $a'',$ i.e., altering the $i$-th element of $\bf a$ and allowing the other ads to follow the rule ${\bf a}_{-i}.$ Then the unit-level effect is quantified via
	$
	\text{UE}_i(a', a'', {\bf a}) = \E\big[ Y_i(a', {\bf a_{-i}}) \big] - \E\big[ Y_i(a'', {\bf a_{-i}}) \big]. 
	$
	
	\item \textit{Spillover effect}: defined as the effect of holding an ad’s  block allocation fixed but modifying the block allocations of other ads on the pageview. 
	Assume we are interested in comparing two allocation rules ${\bf a'}$ and ${\bf a''}$ where the the $i$-th element in each rule is fixed to $a.$ Then the spill-over effect is quantified via 
	$
	\text{SE}_i(a, {\bf a'}, {\bf a''}) = \E\big[ Y_i(a, {\bf a'_{-i}}) \big] - \E\big[ Y_i(a, {\bf a''}_{-i}) \big]. 
	$
	
	\item \textit{Overall effect}: defined as the effect of allocation rule $\bf a$ versus $\bf a'$ on the outcome of the $i$-th ad, which can be quantified via 
	$
	\text{OE}_i({\bf a}, {\bf a'}) = \E\big[ Y_i({\bf a}) \big] - \E\big[ Y_i({\bf a'}) \big]. 
	$
	
	\item \textit{Average overall effect}: defines as a pageview-level comparison of two different allocation rules. This would require an average over all the overall effects	computed on a single pageview, i.e., 
	$
	\text{AOE}({\bf a}, {\bf a'}) = \frac{1}{m}  \ \sum_{i = 1}^m \E\big[ Y_i({\bf a}) \big] - \E\big[ Y_i({\bf a'}) \big]. 
	$
	
\end{enumerate}

\subsection{Identification Assumptions}

Counterfactuals cannot in general be identified from data alone, and require assumptions. It is straightforward to see that all the effects described above involve  counterfactual mean contrasts of the form $\E[Y_i({\bf a})].$ Thus if we can identify this counterfactual mean, all the effects described are identifiable.  In order to identify the counterfactual mean $\E[Y_i({\bf a})],$ we make the following three assumptions: 
(i) \textit{Allocational consistency:}  $Y_i({\bf a}) = Y_i$ if ${\bf A = a}.$ In other words, the potential outcome agrees with the observed outcome when the allocational intervention agrees with the observed allocations,
(ii) \textit{Positivity:} $p({\bf A} = {\bf a} \mid {\bf X} = {\bf x}) > 0, \forall {\bf a} \in \mathfrak{X}_{\bf A}$ and $ \forall {\bf x} \in \mathfrak{X}_{\bf X},$ and 
(ii) \textit{Network conditional ignorability:}  $Y_i({\bf a}) \ci {\bf A} \mid {\bf X}.$ In other words, all the common confounders between each $A_j \in {\bf A}$ and $Y_i$ are measured. 

Given the structural equation model described in (\ref{eq:sem}), the represented causal model in Fig.~\ref{fig:npsem}(a), and the corresponding SWIG in Fig.~\ref{fig:npsem}(b), we can easily verify that network conditional ignorability holds in our model. By rules of d-separation, all the paths from $Y_i({\bf a})$ to each $A_j$ is blocked by conditioning on ${\bf X}.$ Under the aforementioned assumptions, the identifying functional for $\E[Y_i({\bf a})]$ is then obtained as follows,
\begin{align}
\E\big[Y_i({\bf a})\big] = \E\Big[ \E\big[ Y_i \mid {\bf A} = {\bf a}, {\bf X}  \big]  \Big],
\label{eq:target_func}
\end{align}%
where the outer expectation is taken with respect to the marginal distribution over $\bf X$, i.e., $p({\bf X}).$ For a general theory describing when causal inference with interference is possible, interested readers can refer to  \cite{sherman2018identification}.

\subsection{Estimation of Causal Effects}

We set our target of inference to be $\psi = \E[Y_i({\bf a})]$ which is identified via (\ref{eq:target_func}). In order to compute $\psi,$ we use the \textit{augmented inverse probability weighting} (AIPW) estimator, given as
{\small
\begin{align}
\widehat{\psi}_{\text{aipw}} 
&= \frac{1}{N} \sum_{n= 1}^N \ 
\bigg[ \
\frac{ \mathbb I({\bf A}_n = {\bf a}) \times \Big( Y_{in} - \E\big[Y_{in} \mid {\bf A = a}, {\bf X}_{n}; \widehat{\alpha}_y\big] \Big) }{ \prod_{i = 1}^m p(A_{in} = a_i \mid {\bf X}_n; \widehat{\alpha}_a)}  + \E\big[Y_{in} \mid {\bf A = a}, {\bf X}_{n}; \widehat{\alpha}_y\big]
\bigg],
\label{eq:aipw}
\end{align}
}%
where 
$\widehat{\alpha}_y$ and $\widehat{\alpha}_a$ are MLE estimates of the parameters in the outcome regression model $\E[Y \mid {\bf A}, {\bf X}]$ and propensity models $p(A_i \mid {\bf X}),$ respectively. The above estimator is consistent \textit{if and only if} either the propensity scores or the outcome regression models are correctly specified. This property is known as \textit{doubly robust}.  
For a more general discussion of semiparametric doubly robust estimators of average causal effects in presence of unmeasured confounders, see \citet{bhattacharya2020semiparametric}. 

\subsection{Verifying and Learning  Causal Structure}
\label{subsec:FCI}

Throughout the paper, we assumed a known causal structure for the ad placement system. To verify the soundness of our presumed causal structure, we adapt structure learning algorithms to learn the underlying mechanisms that give rise to interference. 
There is a rich literature on model selection from observational data in the context of causal inference with no interference \citep{spirtes2000causation}. 
This includes constraint-based algorithms such as PC \citep{spirtes2000causation, colombo2014order}, score-based algorithms such as GES \citep{chickering2002optimal}, and continuous optimization based algorithms such as the ones in \cite{zheng2018dags, bhattacharya2020discovery}. 
 \cite{bhattacharya2019causal} provided a novel algorithm for model selection when units are related through a network of dependencies that can be modeled using a chain graph \citep{lauritzen96graphical}. However, in our context, dependencies are best modeled using DAGs with hidden variables. There exists (conditional independence) constraint-based algorithms such as \textit{fast causal inference} (FCI) and variations of it, such as GFCI and RFCI, that tackle the model selection problem in the presence of unmeasured confounders.  

Click yields are the primary target of interest. Hence, we adapt the FCI algorithm in order to learn the ``causal parents''  of each $Y_i.$ We do this by performing a pre-processing step on the data, where each row corresponds to the information we collect on a single pageview, in order to account for block-level and cross-block interference. As an example, consider pageviews with three impressed ads where we are interested in finding the causal parents of the outcome in the first positioned ad, i.e., $Y_1.$ We pre-process the data as follows: 
For each row, we evaluate the variables in $X_j$ to zero if $A_j = A_1, $ for $j = 1, 2, 3.$ We call this pre-processed data $\mathcal{D}_1.$ We then evaluate the variables in $X_j$ to zero if $A_j \not= A_1, $ for $j = 2, 3.$ We call this pre-processed data $\mathcal{D}_2.$ We then append $\mathcal{D}_2$ to $\mathcal{D}_1, $ column-wise and pass this data to the FCI algorithm. Additional knowledge, such as causal ordering, can be incorporated in the procedure. The FCI algorithm then returns a partial ancestral graph \citep{zhang2008completeness} as the Markov equivalence class. The partial ancestral graph corresponds to a set of ancestral acyclic directed mixed graphs \citep{richardson2002ancestral} that agree on conditional independence constraints on the observed data distribution. Under standard assumptions, that the true model can be represented via an ancestral graph and faithfulness, (asymptotically) FCI and hence our modification of it returns a Markov equivalence class that contains the true underlying model.


\section{Experiments} 
\label{sec:experiments}

In this section, we illustrate the utility of our formalization of the ad interference problem through three separate experiments using Bing PC traffic: (i) estimating the counterfactual mean under interference as described in Section~\ref{sec:effects}, (ii) identifying causally relevant features through structure learning, and (iii) comparing click prediction models with and without accounting for interference. For training and validation purposes, we used data from the first two weeks of June in 2020. The test data comes from the first two weeks of July in the same year. We use random forest classifiers for fitting the propensity score model and the outcome regression model. We focused on pageviews with at least one observed click and 3 impressed ads. We refer to these as positive pageviews.\footnote{We conducted more tests on other types of pageviews, which are omitted here for brevity.}  

\vspace{0.2cm}
\textbf{Calculation of Interference Effects}

\begin{figure}[!t]
	\begin{center}	
		\begin{subfigure}{.5\textwidth}
			\centering
			\includegraphics[scale=0.64]{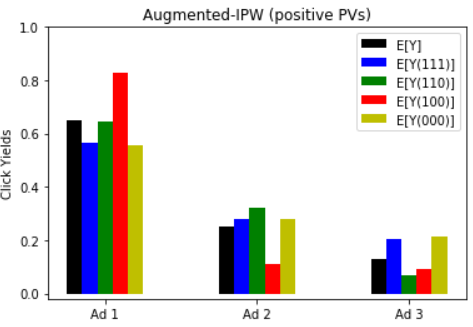}
			\caption{}
		\end{subfigure}%
		\begin{subfigure}{.5\textwidth}
			\centering
			\includegraphics[scale=0.635]{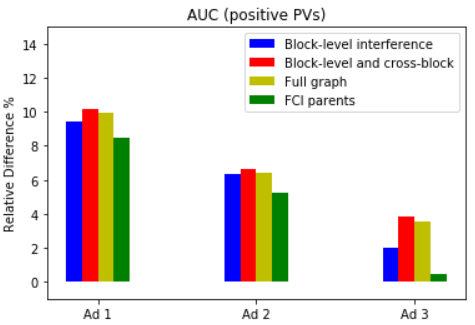}
			\caption{}
		\end{subfigure}%
		\caption{(a) Estimates of the counterfactual mean $\E[Y_i({\bf a})]$ for all possible allocations using AIPW. (b) Relative difference (in percentage) in AUCs with respect to the baseline model.}
		\label{fig:m_aipw}
	\end{center}
\end{figure}

Recall that each allocation rule can be  represented via a binary vector ${\bf a} = (a_1, a_2, a_3).$  As an example, the allocation $(1, 1, 1)$ corresponds to a scenario where all three ads are shown in the Top block. As mentioned in the preliminaries, ads are indexed according to the order in which they appear on the page.  This indexing scheme restricts the state space of all possible allocation rules. 
For instance, an allocation like $(0, 1, 1)$ where the first positioned ad is placed at the Bottom and the rest are on Top is ill-defined and therefore excluded from the set of possible allocation rules. 
%

We use AIPW to estimate the counterfactual mean $\E[Y_i({\bf a})]$ under all possible allocation rules for $\bf a.$ 
The results are shown in Figure~\ref{fig:m_aipw}(a). 
The layout that yields the highest click for each position on the pageview corresponds to the tallest bar on each plot. For instance, the first positioned ad benefits the most from being the sole ad in the Top block, i.e., $\E[Y_1(1, 0, 0)] > \E[Y_1({\bf a})], \forall {\bf a} \neq (1, 0, 0).$ However the corresponding optimal layout for the first positioned ad is not coherent with the optimal layout of other ads. For instance, the second positioned ad benefits the most from being on the Top block as well. On the other hand, the last positioned ad benefits slightly more when all ads are placed at the Bottom. In order to find a coherent optimal layout yielding the highest number of overall clicks, we need to compare the average click response over all positions on the pageview, i.e., the average overall effect $\frac{1}{m} \sum_{i = 1}^m \ \E[Y_i({\bf a})],$ for all possible $\bf a$.

\begin{table*}[!t]
	\begin{center}	
		\caption{Estimated values for the counterfactual mean $\E[Y_i({\bf a})]$ for all possible $\bf a,$ along with the $95\%$ confidence intervals. The observed $\E[Y_i]$ is reported on the last column. }
		\label{tab:m_aipw}
		\scalebox{1}{
			\begin{tabular}{| c | c | c | c | c | c |} 
				\hline 
				{\small } & 
				$\E[Y_i(1, 1, 1)]$ & 
				$\E[Y_i(1, 1, 0)]$ & 
				$\E[Y_i(1, 0, 0)]$ & 
				$\E[Y_i(0, 0, 0)]$ & 
				$\E[Y_i]$  
				 \\  [0.1cm]
				\hline
				First Ad 
				 & $0.57 \ \pm \ 0.006$ & $0.64 \ \pm \ 0.004$ &  $0.83 \ \pm \ 0.004$ & $0.56 \ \pm \ 0.005$ & $0.65$ \\  [0.1cm] 
				Second Ad  & $0.28 \ \pm \ 0.007$ & $0.32 \ \pm \ 0.005$ &  $0.11 \ \pm \ 0.003$ & $0.28 \ \pm \ 0.005$ & $0.25$  \\  [0.1cm] 
				Third Ad  & $0.20 \ \pm \ 0.006$ & $0.07 \ \pm \ 0.002$ &  $0.09 \ \pm \ 0.003$ & $0.21 \ \pm \ 0.005$ & $0.13$ \\  
				[0.1cm] 
				\hline 
			\end{tabular}
		}
	\end{center}
\end{table*}

Estimated values for all the counterfactual means are reported in Table~\ref{tab:m_aipw} along with the corresponding $95\%$ confidence intervals.
We can use this table to compute various effects that were discussed in the previous section. For instance, the following contrast gives us the unit-level effect for $Y_2$ under allocation rule ${\bf a} = (1, 0, 0)$: $\text{UE}_2(1, 0, {\bf a}) = \E[Y_2(1, 1, 0) - Y_2(1, 0, 0)] = 0.32 - 0.11 = 0.21 \ (\pm 0.004).$ The spillover effect under allocation rules ${\bf a} = (1, 0, 0)$ and ${\bf a}' = (1, 1, 1)$ is given by $\text{SE}_{2}(1, {\bf a'}, {\bf a}) = \E[Y_2(1, 1, 1) - Y_2(1, 1, 0)] = 0.28 - 0.32 = - 0.04 \ (\pm 0.006).$
The overall effect of $\bf a$ versus ${\bf a}',$ i.e., $\E[Y_2(1, 1, 1) - Y_2(1, 0, 0)]$ is equal to the sum of UE and SE which is $0.17 \ (\pm 0.007).$ 


\vspace{0.2cm}
\textbf{Learning the Causal Structure using FCI}

In this part of the experiment, we use data to learn the parents of each outcome for 
all ads on the pageview; while allowing for both block-level and cross-block interference. We preprocess the data as described in Section~\ref{subsec:FCI}, and use the  implementation of the FCI algorithm in the Tetrad software\footnote{http://www.phil.cmu.edu/tetrad/about.html}. Independence tests are performed using kernel conditional independence tests \citep{zhang2012kernel} with a significance level of $0.01.$ 
On each pageview, we collect $66$ different features. 
Neither plotting the learned graph nor enlisting all parental sets is relevant to the point we like to deliver here. Our primary objective is to show that for a particular positioned ad, features from other ads on the pageview (not necessarily from the same block even) are directly relevant to the clickability of the ad. In order for us to report the results in a more concise and clear way, we divide the ad-specific features into four distinct categories: (a) Calculated scores, such as \textit{PClick, PDefect, Relevance score, etc}, (b) Decorative features, such as \textit{Twitter information, links, and ratings}, (c) Geometric features, such as \textit{line counts, pixel heights, pixel heights from top of the block}, and (d) match type information. We found out that the parent set of each $Y_i$ contains at least one variable in each category of features from a different ad; providing further evidence for the presence of interference among ads.  
In our extended set of experiments, we learned that Decorative features are more influential on pageviews with higher number of impressed ads.


\vspace{0.2cm}
\textbf{Improvements in Click Prediction}

Given the set of experiments described above, we are more confident to believe that interference \textit{does} exist among ads. This was shown through both finding effects that are away from zero and learning causally relevant features that originate from other ads on the pageview. We now leverage this knowledge to better estimate the click yields. We considered fitting 5 different sets of models: 
(i) {Baseline}: $p(Y_i \mid L, X_i),$
(ii) {Block-level interference}: $p(Y_i \mid L, X_i, \mathbb{I}(A_j = A_i) \times X_j),$
(iii) {Block-level and cross-block interference}: $p(Y_i \mid L, X_i, \mathbb{I}(A_j = A_i) \times X_j, \mathbb{I}(A_k \neq A_i) \times X_k),$
(iv) {Full graph with no block decomposition}: $p(Y_i \mid L, X),$ and
(v) fitting the model using parents in the output of FCI: $p(Y_i \mid \text{pa}(Y_i)).$ 
We report relative improvements in area under the curve over the baseline in Figure~\ref{fig:m_aipw}(b). All methods that account for interference show improvement over the baseline 
over the baseline, demonstrating the utility of our formalization.  
 It is also worth noting that the performance gains are greater for higher positioned ads compared to lower ones.


\section{Conclusions} 
\label{sec:conclusions}

Despite the intuition that ads should not be scrutinized independently of one another, to the best of our knowledge, there has not been a formal analysis of interference in advertisement placement and sponsored search marketing. In this paper, we formalized the interference problem among ads using the language of causal inference and counterfactual reasoning. We proposed a framework to quantify the interference effects by posing a graphical causal model that accounts for potential underlying interference mechanisms. We described several causal effects that might be of interest in ad placement systems and discussed identification assumptions and estimation strategies for computing these effects. 




\bibliographystyle{plainnat}
\bibliography{references}

\end{document}